\documentclass[runningheads]{llncs}

\usepackage[ruled]{algorithm}
\usepackage{algpseudocode}
\usepackage[T1]{fontenc}
\usepackage{amsmath} 
\usepackage{amssymb}  
\usepackage{hyperref}

\hypersetup{
    colorlinks=true,
    linkcolor=blue,
    filecolor=magenta,      
    urlcolor=cyan,
    pdftitle={Overleaf Example},
    pdfpagemode=FullScreen,
    }
\usepackage{siunitx}
\usepackage{threeparttable}
\usepackage{multirow}
\usepackage{makecell}
\usepackage{booktabs}
\usepackage{array}
\usepackage{xcolor}
\newcolumntype{C}[1]{>{\centering\let\newline\\\arraybackslash\hspace{0pt}}m{#1}}
\newcolumntype{L}[1]{>{\raggedright\let\newline\\\arraybackslash\hspace{0pt}}m{#1}}

\newcommand{\mytag}[1]{{\setlength{\fboxsep}{1.5pt}\colorbox{gray!20}{\texttt{#1}}}}
\usepackage{graphicx}
\begin{document}
\title{
From Scanning Guidelines to Action: A Robotic Ultrasound Agent with LLM-Based Reasoning
}
%
\titlerunning{From Scanning Guidelines to Action}
%
\author{Anonymous}
\author{Yuan Bi\inst{1,2}$^{*}$ \and 
Yiping Zhou\inst{1}$^{*}$ \and
Pei Liu\inst{1} \and
Feng Li\inst{1,2} \and
Zhongliang Jiang\inst{3} \and
Nassir Navab\inst{1}}
\authorrunning{Bi et al.}
%
\institute{Chair for Computer-Aided Medical Procedures and Augmented Reality, Technical University of Munich, Munich, Germany \and
Munich Center for Machine Learning, Munich, Germany \and
Medical Intelligence and Robotic Cognition Lab, The University of Hong Kong, Hong Kong, China
}
\begingroup\def\thefootnote{$*$}\footnotetext{These authors contributed equally to this work.}\endgroup
\maketitle              
\begin{abstract}
Robotic ultrasound offers advantages over free-hand scanning, including improved reproducibility and reduced operator dependency. In clinical practice, US acquisition relies heavily on the sonographer’s experience and situational judgment. When transferring this process to robotic systems, such expertise is often encoded explicitly through fixed procedures and task-specific models, yielding pipelines that can be difficult to adapt to new scanning tasks. In this work, we propose a unified framework for autonomous robotic US scanning that leverages a large language model (LLM)–based agent to interpret US scanning guidelines and execute scans by dynamically invoking a set of provided software tools. Instead of encoding fixed scanning procedures, the LLM agent retrieves and reasons over guideline steps from scanning handbooks and adapts its planning decisions based on observations and the current scanning state. This enables the system to handle variable and decision-dependent workflows, such as adjusting scanning strategies, repeating steps, or selecting the appropriate next tool call in response to image quality or anatomical findings. Because the reasoning underlying tool selection is also critical for transparent and trustworthy planning, we further fine tune the LLM agent using a RL based strategy to improve both its reasoning quality and the correctness of tool selection and parameterization, while maintaining robust generalization to unseen guidelines and related tasks. We first validate the approach via verbal execution on 10 ultrasound scanning guidelines, assessing reasoning as well as tool selection and parameterization, and showing the benefit of RL fine tuning. We then demonstrate real world feasibility on robotic scanning of the gallbladder, spine, and kidney. Overall, the framework follows diverse guidelines and enables reliable autonomous scanning across multiple anatomical targets within a unified system.
Code: \url{https://github.com/yuan-12138/RUSSAgent}; Video: \url{https://youtu.be/pfMOc4e2IGA}.

\end{abstract}

\section{Introduction}

Ultrasound (US) is widely used in clinical practice due to its real-time feedback, portability, and lack of ionizing radiation. However, free-hand scanning is highly operator dependent, with image quality varying with the sonographer’s expertise, which limits standardization across operators and sites and can lead to fatigue and work-related musculoskeletal injuries. Robotic US offers a promising alternative by providing precise and repeatable motion for more standardized and reproducible scanning, while integrated force sensing and control can regulate probe–tissue contact to improve safety and support consistent image quality~\cite{bi2024machine,jiang2023robotic}.
\par
Most prior approaches toward autonomous US scanning follow a rule-based design. These systems typically use an external RGB-D camera to reconstruct the patient surface and a robotic arm to manipulate the US probe. Scanning is then guided by anatomy-specific, hand-crafted protocols, often coupled with dedicated US segmentation models.
Most prior approaches to autonomous US scanning are rule based. They typically use an external RGBD camera to reconstruct the patient surface and a robotic arm to manipulate the probe, with scanning guided by anatomy-specific hand-crafted protocols, often combined with dedicated segmentation models. Representative systems include robotic spine US~\cite{yang2021automatic}, autonomous carotid scanning with tailored protocols~\cite{huang2024robot}, and breathing compensation for aortic US~\cite{velikova2024implicit}, as well as additional components such as force-based vertebral localization~\cite{tirindelli2020force} and pulsation-informed artery segmentation~\cite{huang2023motion}. While effective, these rule-based designs are often task-specific and transfer poorly to new anatomies and scanning objectives.

\par
Beyond hand-crafted rule-based designs, robot learning methods have been explored to acquire scanning behaviors from interaction or expert demonstrations. Early work often trained reinforcement learning (RL) agents in simulation, such as RL navigation toward standard spine planes~\cite{li2021autonomous} and VesNetRL for carotid standard views~\cite{bi2022vesnet}. More recent studies increasingly learn from expert demonstrations to mitigate sim-to-real gaps, including inverse RL to learn the “language of sonography”~\cite{jiang2024intelligent} and Diffusion Policy for robotic carotid scanning~\cite{chen2025ultradp}. While effective in targeted scenarios, achieving flexible behavior and robust generalization across diverse scanning tasks remains challenging.

\par
Large language models (LLMs) have demonstrated strong general-purpose capabilities and can benefit robotics by enabling flexible task understanding and high-level decision making~\cite{raptis2025agentic}. In navigation and manipulation, LLMs are often integrated in a zero-shot manner as high-level planners for tool selection and long-horizon reasoning, while low-level controllers and perception modules execute the actions~\cite{shah2023lm,singh2025malmm,rana2023sayplan,stone2023open}. However, this transfer is less direct in the medical domain, where general-purpose LLMs may lack domain-specific knowledge.
As a result, early attempts to integrate LLMs into robotic US have incorporated clinical scanning guidelines to appropriately constrain the planning process and improve reliability~\cite{xu2024transforming}. Chen~\emph{et al.} \cite{chen2025uspilot} proposed USPilot, an LLM-based framework for robotic US scanning across multiple organs. However, the reasoning capability of the LLM agent is often not explicitly modeled or evaluated in prior work. As a result, existing systems typically emphasize guideline execution and tool use, while offering less insight into the agent’s intermediate reasoning that underlies its planning and tool selection decisions.

\par
In this work, we propose a unified agentic framework for autonomous robotic US, in which an LLM interprets scanning guidelines and translates them into actions by dynamically selecting from a set of perception and control tools. By retrieving protocol steps from handbooks and reasoning over the current scanning state and observations, the agent can execute variable, decision-dependent workflows (e.g., repeating steps, adapting strategies, or adjusting force based on image quality) rather than following a fixed sequence. To strengthen reasoning over planning decisions, we further introduce a RL strategy to fine-tune the LLM agent for more reliable reasoning and tool-calling behavior. We evaluate the proposed framework through verbal interaction experiments on unseen US scanning protocols, and we further demonstrate end-to-end feasibility in real-world robotic experiments on human volunteers across three representative examinations: gallbladder, spine, and kidney.

\begin{figure}[h]
    \centering
    \includegraphics[width=1\textwidth]{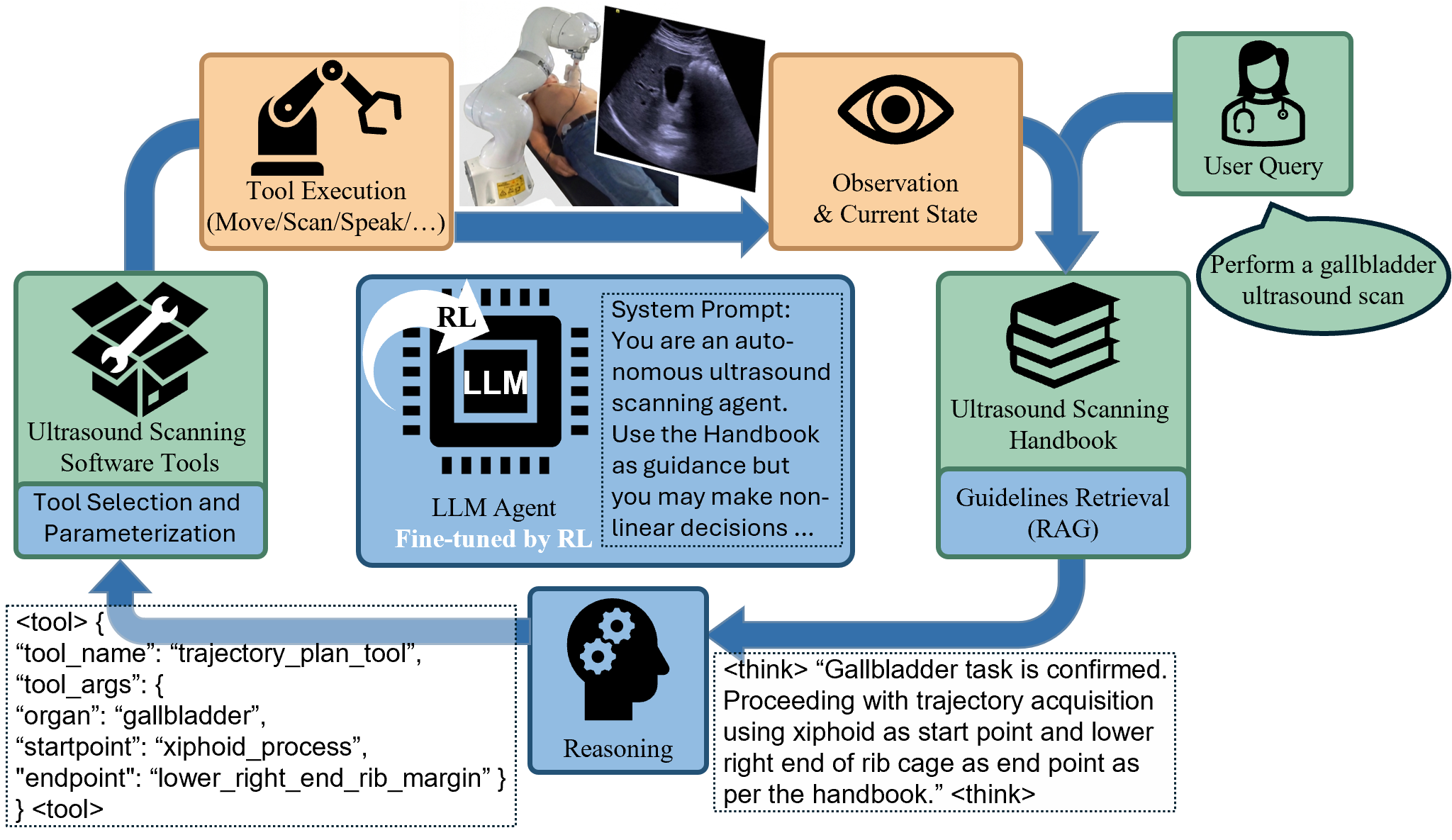}
    \caption{Overview of the proposed guideline-driven agentic robotic US system. An RL fine tuned LLM retrieves guideline steps from the scanning handbook, reasons over the current state and observations, and generates structured tool calls with parameters. Executed tools update the state, enabling closed loop, non linear scanning workflows.}
    \label{fig:pipeline}
\end{figure}

\section{Methods}
LLMs have strong reasoning and decision-making capabilities. We leverage an LLM as a high-level agent for autonomous US scanning by equipping it with software tools for image processing, perception, and robot control, and by providing access to a database of scanning guidelines to supply domain knowledge. Conditioned on the user query and current observations, the agent retrieves guideline steps, selects and parameterizes tools, and executes the requested scanning task. We further apply RL-based fine tuning to adapt a small LLM to robotic US, improving planning-oriented reasoning and tool calling while still generalizing robustly to unseen guidelines and producing plausible reasoning traces to support transparency and clinician inspection.

\subsection{Agentic LLM for Robotic US}
Our proposed framework follows an agentic paradigm similar to \cite{schick2023toolformer,fallahpour2025medrax,daghyani2025echoagent}, but targets autonomous robotic US scanning through guideline-driven planning and tool use rather than verbal or image analysis tasks. It consists of three components: (1) robotic and perception tools for path generation, patient-robot interaction, etc.; (2) a guideline database containing US workflow steps; and (3) an LLM that serves as the high-level planner.
At initialization, a prefix prompt defines the LLM role for sequential problem solving and tool-based response generation. Given a user query, the system retrieves relevant guideline instructions via retrieval-augmented generation~\cite{lewis2020retrieval} and the agent selects and parameterizes tools accordingly. At each step, the agent reasons over the current observations and outputs a \mytag{<think>} reasoning trace followed by a \mytag{<tool>} token with a structured JSON tool call. Tool outputs are appended to the history to guide subsequent decisions, enabling variable, decision-dependent workflows such as repeating steps, adapting scanning strategies, and adjusting trajectory or force based on US observations until the guideline objectives are met.

\subsection{Tools Design}
In order to give the LLM the ability to perform automatic US scanning and analyses the scanning status, specific tools are designed for imaging processing and robot execution.

\begin{figure}[h]
    \centering
    \includegraphics[width=1\textwidth]{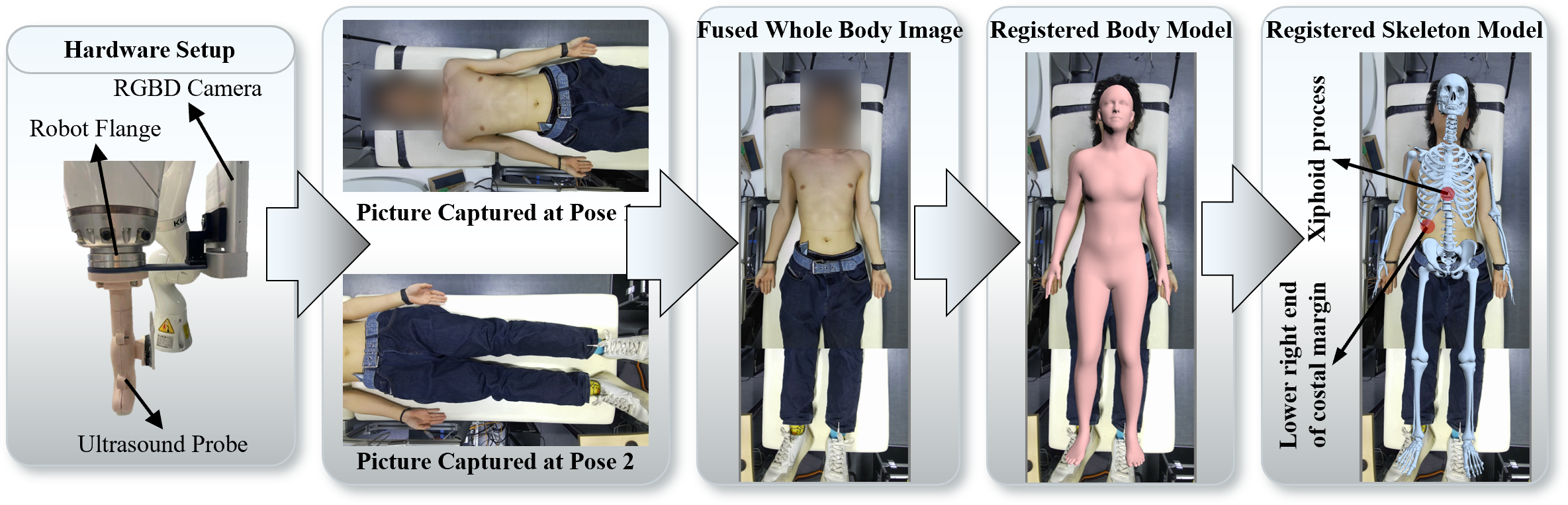}
    \caption{Trajectory planning pipeline for initial scan generation.}
    \label{fig:traj_plan}
\end{figure}

\par
\noindent\textbf{Trajectory Planning Tool} is designed to generate an initial US scanning trajectory based on the input query. As shown in Fig.~\ref{fig:traj_plan}, first, the RGBD camera mounted on the robot end effector is used to capture the patient body surface. Due to the limited field of view of the RGBD camera, multiple views are acquired and fused to reconstruct a whole body surface representation of the patient. A skeletal atlas is then registered to the patient using an open-source model~\cite{li2022cliff,keller2023skin}, which enables the detection of anatomical landmarks for trajectory planning. Based on LLM-provided inputs specifying the target organ and the start and end points of the scan defined with respect to anatomical landmarks, the tool determines a scanning path and projects it into 3D space using the patient point cloud. In summary, the inputs to the tool include the target organ and landmark-referenced start and end points of the scanning trajectory, and the output is a 3D US scanning trajectory.

\par
\noindent\textbf{Robot Execution Tool} receives either a scanning trajectory or a target pose as input, together with the execution speed. It then executes the corresponding motion command to move the robot to the specified pose or along the specified trajectory for US scanning or RGBD image acquisition.

\par
\noindent\textbf{Contact Adjustment Tool} evaluates US image quality and probe contact quality based on the US confidence map~\cite{karamalis2012ultrasound}, and then adjusts the robot pose to establish proper contact with the patient surface~\cite{chatelain2017confidence,bi2025gaze}. The input to this tool is the US image, and the output is adjusted robot pose or trajectory.

\par
\noindent\textbf{Voice Guidance Tool} generates the verbal instructions needed to guide the patient during scanning. These instructions include breath-hold cues and body pose adjustment guidance. 
US scanning is a highly interactive procedure, and effective communication with the patient is essential for adjusting body posture and obtaining an appropriate acoustic window.
For example, such guidance is particularly important for gallbladder scanning, because without proper instructions, such as a deep-inhalation breath hold, it is often difficult to visualize the gallbladder beneath the rib cage and avoid acoustic shadowing caused by bone. The input to the tool is the message that needs to be conveyed to the patient.

\par
\noindent\textbf{Trajectory Refinement Tool} evaluates the results of the initial scan and refines the trajectory when necessary. Since the initial scan is planned only based on the registered skeletal atlas and anatomical priors, the target organ may not appear in the center of the US image. Therefore, after the initial scan, a target-organ-specific segmentation model is applied to localize the organ in 3D space, and the scanning trajectory is then replanned to enable a refined scan. If the organ is already sufficiently centered and well visualized in the initial scan, no refinement is performed. The input to the tool is the tracked US data from the initial scan, and the output is a replanned trajectory when it is necessary.

\subsection{RL-based Fine-tuning}
Given the response time constraints in robotic applications, using a very large LLM is often impractical. Although the proposed framework is not a vision language action model that requires a response frequency above 20 Hz, an agentic system for robotic US still benefits from sufficiently fast response time on accessible GPU hardware. Therefore, fine tuning a smaller model is necessary to obtain reliable reasoning and tool calling behavior, since smaller models typically have more limited capacity and generalization than larger models. At the same time, the fine tuned model should still robustly generalize to other guidelines and related tasks.
\par
To this end, we construct a guideline driven tool use dataset by collecting stepwise agent traces from a set of training US scanning guidelines. For each guideline step, a larger LLM is used to generate a reference response consisting of \mytag{<think>} reasoning and a \mytag{<tool>} JSON command with the correct tool and parameters. We first perform supervised fine tuning (SFT) on these reference traces to teach the small model the structured output format and basic tool use behavior. We then apply RL, which is increasingly used to improve reasoning behavior and response correctness in LLMs~\cite{guo2025deepseek}. Specifically, we adopt Proximal Policy Optimization to further optimize the agent. 
We define a dense reward for each predicted tool call by comparing the structured output
$\hat{y}=\{\hat{t},\hat{\mathbf a}\}$ to a reference tool call $y=\{t,\mathbf a\}$, where
$t$ and $\hat{t}$ are tool names and $\mathbf a,\hat{\mathbf a}$ are JSON argument dictionaries.
Let $\mathcal A$ and $\hat{\mathcal A}$ denote the sets of argument keys in $\mathbf a$ and $\hat{\mathbf a}$, and let
$\mathcal K \subseteq \mathcal A$ be the set of scored keys with tool-specific weights $w_k$.

\begin{equation}
r(\hat{y},y)=
\begin{cases}
-0.5, & \hat{t}\neq t \\[2pt]
\mathrm{clip}_{[0,1]}\!\left(0.1+0.9\, s_{\mathrm{args}}\right), & \hat{t}=t
\end{cases}
\label{eq:reward}
\end{equation}
This equation assigns negative reward if the predicted tool name is incorrect, and otherwise combines a small base reward with a dense argument score.

\begin{equation}
s_{\mathrm{args}}=
\mathrm{clip}_{[0,1]}\!\left(
0.5\, s_{\mathrm{pres}} + 0.5\, s_{\mathrm{corr}} - 0.1\, s_{\mathrm{extra}}
\right)
\label{eq:args_score}
\end{equation}
This equation defines the dense argument score as a mixture of key presence and value correctness, with a penalty for spurious extra arguments.

\begin{equation}
s_{\mathrm{pres}}
=
\frac{\sum_{k\in\mathcal K} w_k\, \mathbf{1}\!\left[k\in\hat{\mathcal A}\right]}
{\sum_{k\in\mathcal K} w_k},
\qquad
s_{\mathrm{corr}}
=
\frac{\sum_{k\in\mathcal K} w_k\, \mathbf{1}\!\left[\hat{a}_k \approx a_k\right]}
{\sum_{k\in\mathcal K} w_k}
\label{eq:pres_corr}
\end{equation}
Here, $s_{\mathrm{pres}}$ rewards including the required argument keys, while $s_{\mathrm{corr}}$ rewards correct argument values, using exact matching for discrete fields and tolerance-based matching for selected numeric fields in the relation $\hat{a}_k \approx a_k$.

\begin{equation}
s_{\mathrm{extra}}
=
\frac{\left|\hat{\mathcal A}\setminus \mathcal A\right|}
{\max\left\{1,\;|\hat{\mathcal A}|\right\}} .
\label{eq:extra}
\end{equation}
This term penalizes extra argument keys that are not expected by the reference tool call, encouraging concise and executable tool commands.
The model is fine tuned with LoRA~\cite{hu2022lora} using a rank of 16 for SFT and a rank of 8 for RL.

\section{Experiments and Results}

\noindent\textbf{Verbal Interaction Experiments}
We create 20 US scanning guidelines and split them into 10 training guidelines and 10 held-out testing guidelines, with no overlap in guideline content. The tool set remains identical across training and testing. To obtain reference outputs, we use Qwen3-30B-A3B-Instruct-2507~\cite{qwen3technicalreport} to generate guideline-conditioned agent traces consisting of \mytag{<think>} reasoning and \mytag{<tool>} JSON commands. We fine-tune Qwen3-4B-Instruct-2507~\cite{qwen3technicalreport} on the training guidelines and evaluate on the held-out guidelines to assess robustness to unseen guideline instructions, measuring tool selection and parameterization accuracy together with structured output validity.

\par
We then compare the base model and the RL fine tuned model under identical guideline and tool settings. Specifically, we report the overall scanning task success rate, defined as successful completion of the guideline specified scan. In addition, we evaluate step-level tool use by measuring, at each guideline step, whether the agent selects the correct tool and provides correct tool arguments. The results are shown in Tab.~\ref{tab:perf_comparison}. Overall, RL fine tuning leads to a clear improvement across both task success and step-level tool calling accuracy.

\begin{table}[h]
\centering
\caption{Performance comparison before and after fine-tuning.}
\label{tab:perf_comparison}
\begin{tabular}{lc|c}
\toprule
 & Base Model & After Fine Tuning \\
\midrule
Step-wise Accuracy   & 0.6512 & 0.8973 \\
Overall Success Rate & 0.5384 & 0.9230 \\
\bottomrule
\end{tabular}
\end{table}

\begin{figure}[h]
    \centering
    \includegraphics[width=1\textwidth]{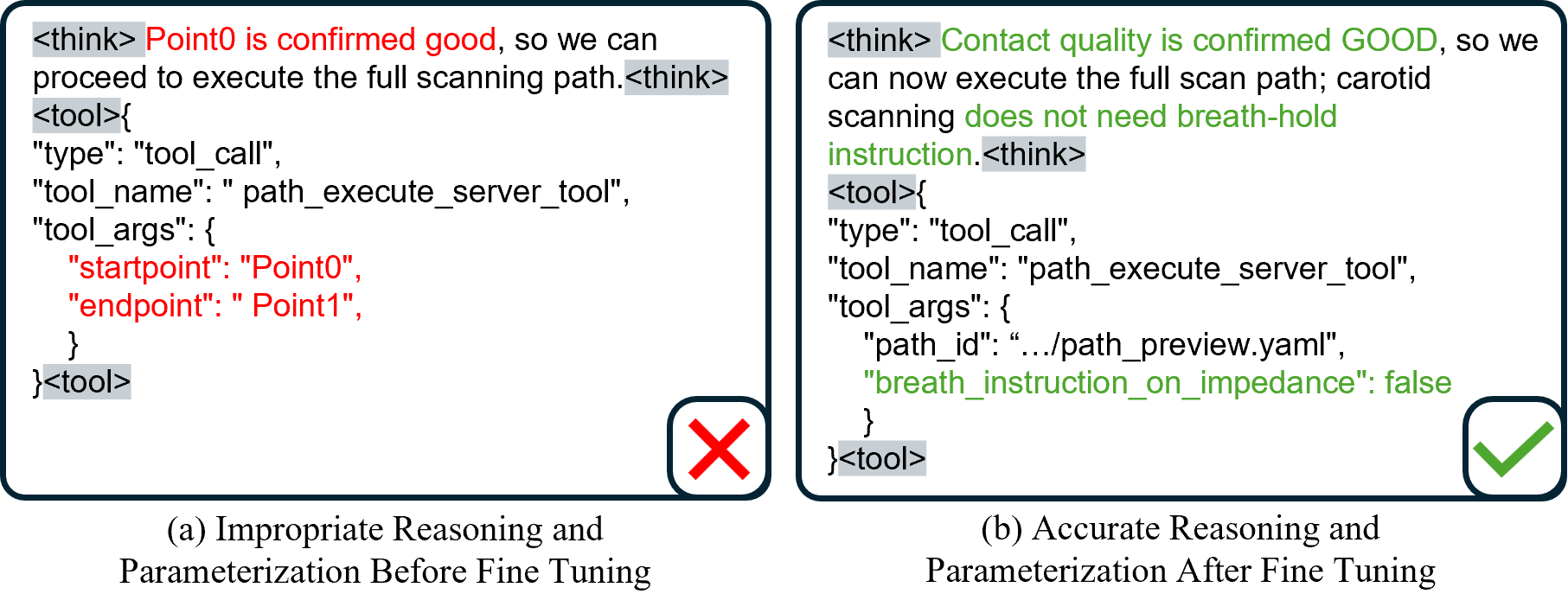}
    \caption{Verbal Experiments Examples.}
    \label{fig:verbal_test}
\end{figure}

\par
We also include qualitative examples comparing tool calls generated by the base model and the RL fine tuned model (Fig.~\ref{fig:verbal_test}), illustrating differences in tool parameterization and reasoning at representative guideline step.
Qualitatively, the base model often produces minimal or uninformative reasoning traces, whereas the RL fine tuned model generates more coherent and task-relevant rationales alongside its tool calls.

\begin{figure}[h]
    \centering
    \includegraphics[width=1\textwidth]{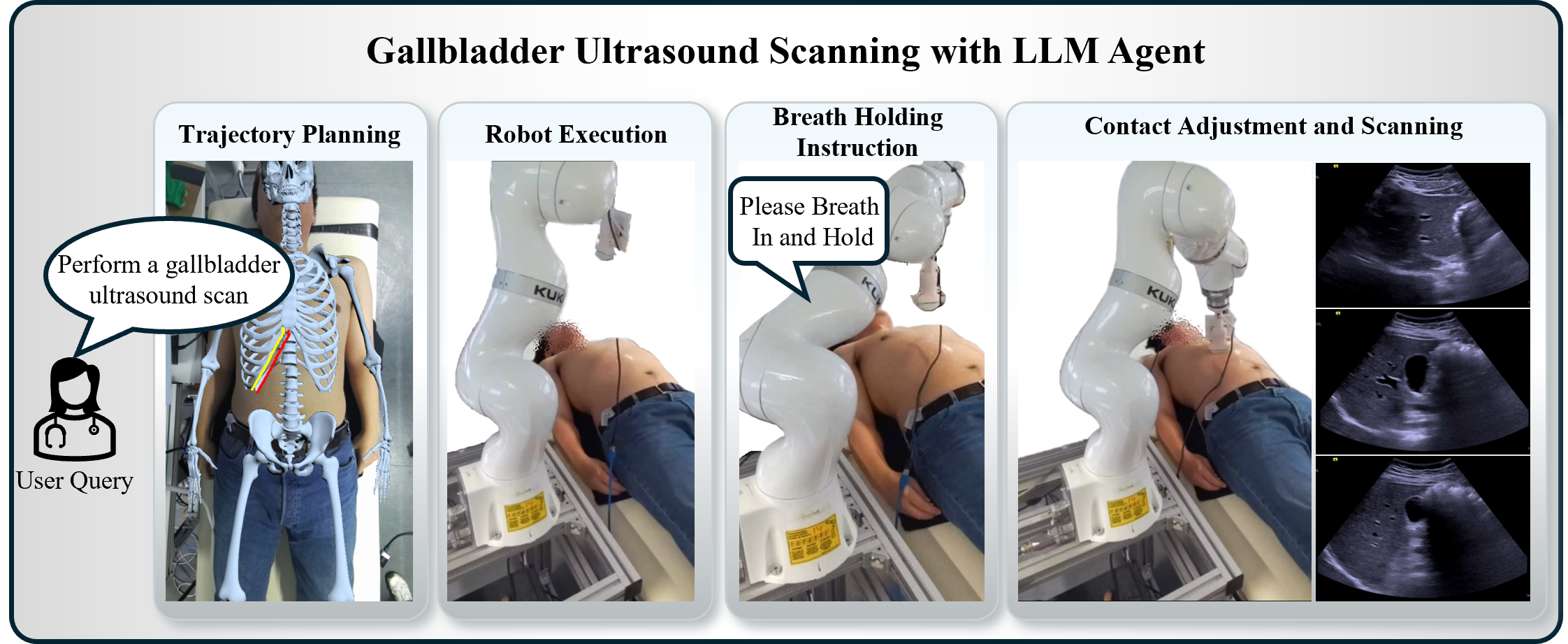}
    \caption{Real Scenario Experiments Examples.}
    \label{fig:real_test}
\end{figure}

\noindent\textbf{Real-world Robotic Experiments}
We further validate the proposed system in real-world robotic US experiments on three human volunteers, demonstrating autonomous scanning of the gallbladder, spine, and kidney. Ethical approval was obtained from the institutional review board of the anonymised institute, and all participants provided written informed consent.
The scanning setup is shown in Fig.~\ref{fig:traj_plan}. We use a KUKA LBR iiwa robot equipped with a Siemens ACUSON Juniper US system and a 5C1 probe mounted via a 3D printed holder, with US images acquired through an Epiphan frame grabber. An RGBD camera, Azure Kinect by Microsoft, is mounted on the robot end effector to capture the 3D surface of the scanning target.
We evaluate task success based on whether the system can correctly visualize the target organ in the US image. One exemplary scanning scenario is shown in Fig.~\ref{fig:real_test}
All spine and kidney experiments were successful. For gallbladder scanning, one case failed due to the higher procedural complexity, which requires breath hold instructions and close coordination between the patient and the robot.

\par

\section{Conclusion}
In this work, we presented a guideline-driven agentic framework for autonomous robotic ultrasound that uses an RL fine tuned LLM to retrieve scanning guidelines, reason over the current scanning state, and generate structured tool calls for execution. Experiments on held-out guidelines and real-world volunteer studies across gallbladder, spine, and kidney scanning demonstrate that the proposed system can robustly follow diverse guidelines and achieve reliable autonomous scanning within a unified pipeline.




\bibliographystyle{splncs04}
\bibliography{references}

\end{document}